\documentclass[conference]{IEEEtran}

\usepackage[normalem]{ulem}
\usepackage{algorithm}
\usepackage[noend]{algpseudocode}
\usepackage{float}
\usepackage{booktabs, makecell, tabularx}
\usepackage{comment}
\usepackage{amsmath}
\setcellgapes{3pt}
\usepackage{amsfonts}
\usepackage{xcolor}
\usepackage{soul}
\usepackage{multirow}
\usepackage{arydshln}
\usepackage{bookmark}
\usepackage{adjustbox}
\usepackage{tikz}
\usetikzlibrary{shapes.geometric, arrows, positioning}
\usepackage{graphicx}
\usepackage{caption}
\usepackage{subcaption}
\usepackage{amssymb}
\usepackage{textcomp}
\usepackage{listings}
\usepackage{tabularray}
\usepackage{cite}
\usepackage{siunitx}
\usepackage{microtype}
\newcommand{\shrink}[1][1]{\vspace{-#1\dimexpr0.25cm\relax}}

\definecolor{listinggreen}{rgb}{0,0.6,0}
\definecolor{listinggray}{rgb}{0.5,0.5,0.5}
\definecolor{listingmauve}{rgb}{0.58,0,0.82}
\definecolor{listingkeywordcolor}{rgb}{1.0,0.4,0.0}
\definecolor{listinglightgray}{rgb}{0.8863,0.8863,0.8863}

\newcommand{\CodeSize}{\footnotesize}
\newcommand{\SmallCodeSize}{\scriptsize}

\newcommand{\NbNodes}{16}

\newcommand{\toolname}{MLIR RL}

\lstdefinelanguage{MLIR}{
  morekeywords={
    module,
    func.func,
    tensor,
    affine_map,
    affine.apply,
    scf.forall,
    scf.forall.in_parallel,
    linalg.matmul,
    tensor.extract_slice,
    tensor.parallel_insert_slice,
    vector.transfer_read,
    vector.transfer_write,
    vector.multi_reduction,
    arith.constant,
    arith.mulf,
    arith.addf,
    arith.subi,
    call,
    return,
    ins,
    outs
  },
  sensitive=true,
  morecomment=[l]{//},
  morestring=[b]"
}

\makeatletter
\DeclareTextCommand{\textquotedbl}{OT1}{\char`\"}
\makeatother

\makeatletter
\newcommand{\linebreakand}{%
  \end{@IEEEauthorhalign}
  \hfill\mbox{}\par
  \mbox{}\hfill\begin{@IEEEauthorhalign}
}
\makeatother 

\begin{document}

\title{A Reinforcement Learning Environment for Automatic Code Optimization in the MLIR Compiler}

\author{

\IEEEauthorblockN{Mohammed Tirichine}
\IEEEauthorblockA{\textit{New York University Abu Dhabi}\\
Abu Dhabi, UAE\\
\textit{ESI}\\
Algiers, Algeria\\
tirichine.m@nyu.edu}

\and

\IEEEauthorblockN{Nassim Ameur}
\IEEEauthorblockA{\textit{New York University Abu Dhabi}\\
Abu Dhabi, UAE\\
\textit{ESI}\\
Algiers, Algeria\\
na3758@nyu.edu}

\and

\IEEEauthorblockN{Nazim Bendib}
\IEEEauthorblockA{\textit{New York University Abu Dhabi}\\
Abu Dhabi, UAE\\
\textit{ESI}\\
Algiers, Algeria\\
nb3891@nyu.edu}

\vspace{0.3cm}
\linebreakand

\IEEEauthorblockN{Iheb Nassim Aouadj}
\IEEEauthorblockA{\textit{New York University Abu Dhabi}\\
Abu Dhabi, UAE\\
ia2280@nyu.edu}

\and

\IEEEauthorblockN{Djad Bouchama}
\IEEEauthorblockA{\textit{New York University Abu Dhabi} \\
Abu Dhabi, UAE\\
\textit{USTHB}\\
Algiers, Algeria\\
db5394@nyu.edu}

\and

\IEEEauthorblockN{Rafik Bouloudene}
\IEEEauthorblockA{\textit{New York University Abu Dhabi} \\
Abu Dhabi, UAE\\
\textit{USTHB}\\
Algiers, Algeria\\
rb5953@nyu.edu}

\vspace{0.3cm}
\linebreakand

\IEEEauthorblockN{\centering Riyadh Baghdadi}
\IEEEauthorblockA{\textit{\centering New York University Abu Dhabi}\\
Abu Dhabi, UAE\\
baghdadi@nyu.edu}

}

\maketitle
\pagestyle{plain}

\begin{abstract}
Code optimization is a crucial task that aims to enhance code performance. However, this process is often tedious and complex, highlighting the necessity for automatic code optimization techniques. Reinforcement Learning (RL) has emerged as a promising approach for tackling such complex optimization problems. In this project, we introduce \toolname{}, an RL environment for the MLIR compiler, dedicated to facilitating MLIR compiler research and enabling automatic code optimization. We propose a multi-discrete formulation of the action space where the action space is the Cartesian product of simpler action subspaces. We also propose a new method, called level pointers, to reduce the size of the action space related to the loop interchange transformation. This enables more efficient and effective learning of the policy.
To demonstrate the effectiveness of \toolname{}, we train an RL agent to optimize MLIR Linalg code, targeting CPU. The code is generated from two domain-specific frameworks: deep-learning models generated from PyTorch, and LQCD (Lattice Quantum Chromodynamics) code generated from an LQCD compiler.
The result of this work is a research environment that allows the community to experiment with novel ideas in RL-driven loop-nest optimization.

\end{abstract}


\thispagestyle{plain}

\section{Introduction}

Writing high-performance code for compute-intensive applications is a challenging task that requires significant expertise. While manual code optimization can reduce execution time significantly, it is time-consuming and error-prone, making automatic compiler code optimization increasingly important.

Several techniques have been proposed to enable automatic code optimization in compilers, including the use of integer linear programming (ILP) to find the best code optimizations~\cite{bondhugula2008practical,sven2013}, and using tree-search methods guided by deep learning cost models~\cite{adams2019halide,zheng2020ansor,baghdadi2021deep}.

More recent work has explored the use of Reinforcement Learning (RL) for automatic code optimization~\cite{pecenin2019optimization,chameleon,pearl,rl_hsl, autophase2019, huanting2022ss}. 
These approaches model the problem as sequential decision-making, where an RL agent selects a sequence of compiler transformations.
While such approaches have made progress, they still have limitations: some are semi‑automatic (requiring user‑specified directives), others are implemented in research compilers with limited support for complex workloads and code transformations, and many target narrow domains (e.g., DNN graphs only), limiting generality. 

Making progress in this area requires the development of novel RL environments integrated within robust compilers that expose general, parameterized loop transformations across domains.
In this context, the MLIR (Multi‑Level Intermediate Representation) compiler~\cite{mlir} has emerged as a strong fit.
It is an industrial-grade compiler infrastructure with a robust implementation and widespread adoption in industry and academia. Its multi-dialect architecture exposes a rich set of loop and data-layout transformations that are fine-grained and composable. Its ecosystem spans a wide range of front ends and back ends, enabling a single setup to cover diverse languages and hardware targets. These strengths make MLIR a strong foundation for research on RL-based automatic code optimization.
However, to our knowledge, there is currently no RL environment for automatic code optimization natively integrated with MLIR.

Although prior work, such as CompilerGym~\cite{cummins2022compilergym}, provides an RL environment for LLVM, it is not directly usable in MLIR without substantial adaptation. In particular, building an MLIR RL environment requires solving many challenges. 
First, the action space modeling in CompilerGym is not suitable for automatic code optimization: CompilerGym models actions as selecting from a fixed set of LLVM passes (i.e., pass ordering over LLVM IR). In MLIR, loop‑level optimization requires parameterized, location‑specific transformations (e.g., choosing loops on which a transformation is applied, choosing tiling factors, fusion targets, etc.). Because CompilerGym does not expose such a structured, parameterized action space or enumerate legal application sites, MLIR optimization cannot be easily supported without effectively building a new environment.
The second challenge when building an RL environment for automatic code optimization is the large size of the action space: each transformation can be applied at different loop levels and has multiple parameters, yielding a combinatorial number of choices.
Third, implementing this action space is nontrivial: MLIR’s multi‑dialect architecture exposes fine‑grained, parameterized transformations, and building an action space that correctly applies these transformations requires significant engineering effort.
The fourth challenge is the high cost of training an agent for the RL environment: evaluating long transformation sequences across many programs and repeatedly executing each variant to obtain stable measurements is computationally expensive, especially for large inputs.
During development, this end‑to‑end training must be repeated many times to validate design choices, compare agents, and tune hyperparameters, making the cumulative training cost substantial. Such repeated training demands computing resources that are sometimes beyond the reach of the research community.
These challenges make it difficult to simply adapt existing RL environments, motivating the need to build a specialized RL environment for automatic code optimization in MLIR.

To address this gap, we propose \toolname{}, an RL environment for automatic loop‑level code optimization in MLIR.
We propose an RL agent with a multi-discrete action space: it first selects the transformation to apply and then its parameters.
This is done by formulating the action space as the Cartesian product of much smaller subspaces. This enables the RL agent to learn a better policy and potentially find better sequences of optimizations. To further reduce the size of the action space, we propose a novel policy network that uses level pointers for the loop interchange transformation: instead of enumerating all the possible loop interchanges (which would be high), the network decides about the loop order, level by level, one at a time, significantly reducing the size of the action space and improving policy learning.

To demonstrate the effectiveness of \toolname{}, we train an RL agent to optimize MLIR code generated from two domain-specific frameworks: deep-learning models generated from PyTorch, and LQCD (Lattice Quantum Chromodynamics) code generated from an LQCD compiler. LQCD is a branch of theoretical nuclear physics that uses large-scale simulations of quarks and gluons on a spacetime lattice to study the strong force and the structure of matter. MLIR RL specifically optimizes code in the Linalg MLIR dialect targeting CPU (Linalg is an MLIR IR dialect designed for linear algebra and deep learning computations).

We evaluate \toolname{} against state-of-the-art frameworks (PyTorch~\cite{pytorch}, PyTorch compiler~\cite{pytorchjit}, Halide autoscheduler~\cite{mullapudi2016automatically}, and Halide RL~\cite{pecenin2019optimization}).
The outcome of this work is a research infrastructure that the community can use to explore novel ideas in the topic of RL-based loop nest optimization.

In summary, the contributions of this paper are:

\begin{enumerate}
    \item We propose \toolname{}, an RL environment for automatic code optimization in the MLIR compiler, facilitating further research in this area.
    \item We propose a multi-discrete action space and use level pointers to reduce the size of the action space and improve policy learning.
    \item We implement and evaluate our proposed approach, showing its effectiveness in optimizing MLIR code.
    \item We release \toolname{} to the community\footnote{\url{https://github.com/Modern-Compilers-Lab/MLIR-RL}}.
\end{enumerate}

The rest of the paper is structured as follows: we begin with a background on MLIR and RL in compilers, followed by a detailed description of our RL environment, including the action space, states, and rewards. We then present the multi-discrete policy network and the use of level pointers. Finally, we evaluate \toolname{} on a set of benchmarks and compare it to four state-of-the-art compilers and frameworks: PyTorch, the PyTorch compiler, Halide autoscheduler, and Halide RL.

\section{Background and Related Work}

Many efforts in the compiler community have focused on automatic code optimization~\cite{Iri88,feautrier_array_1988,wolf1991loop,lefebvre_automatic_1998,Qui00,thies_unified_2001,Darte_contraction_2005,bondhugula2008practical,baghdadi2015PhD,baghdadi2019tiramisu,baghdadi2018tiramisu1,trifunovic_graphite_2010,polly,tobias_hexagonal_cgo13,Vasilache2018TensorCF,baghdadi2011speculation,pouchet.11.popl,baghdadi2020tiramisuDNNDenseSparse,adams2019halide,baghdadi2021deep,merouani2020deep,chen2018learning,zheng2020ansor,mendis2019ithemal,polygym,merouani2024looper,hakimi2023hybrid,mezdour2023deep,bendib2024reinforcement,autophase2019,mlgo,cummins2022compilergym,felix,looper_data_arxiv25,predictivepolyhedral,huanting2022ss,rl_hsl}. In this section, we provide an overview of relevant background and related work, focusing on MLIR and the use of Reinforcement Learning (RL) in compiler optimization. 

\subsection{MLIR}

MLIR \cite{mlir} is a framework designed to address the needs of modern compilers and heterogeneous hardware. It provides a flexible infrastructure for defining multiple levels of intermediate representations through various dialects, facilitating code translation and optimization across diverse domains. Notably, MLIR includes the Linalg dialect for linear algebra operations, the Affine dialect for expressing affine loops, and the Vector dialect for vectorization. MLIR also improves code optimization in DNN frameworks such as TensorFlow and PyTorch, offering a unified infrastructure that improves performance across frameworks.

\subsection{Machine Learning for Compilers}

Machine learning has been used to improve compiler optimizations, notably to train a cost model that estimates the performance of optimized code. It was used in Tiramisu~\cite{baghdadi2021deep,merouani2024looper}, Halide~\cite{adams2019halide}, and TVM~\cite{TVM} to empower search algorithms such as beam search to efficiently find better schedules. Because these compilers rely on tree-search algorithms, they need to explore a large number of optimization candidates to find the best sequence of code optimizations. The Halide autoscheduler, for instance, explores millions of schedule candidates~\cite{adams2019halide}, which makes the process of code optimization slow. To constrain the search space, these compilers often impose restrictions, such as Tiramisu's fixed exploration order~\cite{baghdadi2021deep}. Due to these limitations, recent efforts have shifted toward RL as a more scalable solution for automatic code optimization.

\subsection{Reinforcement Learning for Compilers}

More recent work has increasingly focused on RL due to its potential to train systems that automatically select the best sequence of actions, which is highly relevant for tasks such as loop optimization. For instance, Halide RL \cite{pecenin2019optimization} employed RL to determine the best sequence of code optimizations and their parameters to minimize the execution time of image processing pipelines. Halide RL is not fully automatic, though, as it requires an initial input set of directives provided by a user for the RL agent to select from. This is different from our proposed RL environment, where the whole optimization process is automated.

Other work \cite{rl_hsl, autophase2019} utilized RL to target the problem of phase ordering. It aims to automate high-level synthesis (HLS) by selecting the best order of compiler passes. Our proposed RL approach, rather than relying on passes, tackles the task of code optimization by selecting the optimizations to apply, their parameters, in which order to apply them, and on which part of the code. In addition, AutoPhase \cite{autophase2019} targets HLS and does not target CPUs, which we focus on.

Prior works like Chameleon~\cite{chameleon}, REGAL~\cite{paliwal2020reinforced}, and X-RLflow~\cite{he2023xrlflowgraphreinforcementlearning} leverage RL to accelerate DNN graphs, focusing on high-level tasks such as parameter tuning for convolutions, model parallelism, or graph rewriting. In contrast, our method operates at a lower level (MLIR linear algebra dialect) rather than on deep learning graphs. This allows our RL environment to target multiple domains, which we demonstrate using both deep learning models and LQCD computations.

SuperSonic~\cite{huanting2022ss} introduces a meta-optimizer to search and tune RL architectures, providing a tool for the automatic design of RL environments. It addresses a problem that is orthogonal to our contribution, and our RL environment can also benefit from these techniques.

To facilitate research, CompilerGym~\cite{cummins2022compilergym} provides RL environments for tasks like LLVM phase ordering and GCC flag selection. Similarly, PolyGym~\cite{polygym} leverages polyhedral optimization for general-purpose computation.

While these two environments are important milestones towards democratizing research on RL in compilers, they are not designed for automatic code optimization in the MLIR compiler. Therefore, one needs to spend a significant effort to fully integrate them into MLIR before being able to use them in their research. Moreover, adapting them to the task of automatic code optimization in MLIR is tedious. This is mainly because developing an effective action space that has a comprehensive list of optimizations in the MLIR compiler requires effort and expertise. Most of our effort in building our proposed RL environment was spent on building this effective action space. 

The MLIR compiler is now widely used by the research community and is becoming the backbone of several deep learning frameworks and domain-specific language compilers. Therefore, we believe that a specialized RL environment specifically designed for automatic code optimization in MLIR is needed.

\subsection{MLIR-based Compilers for Machine Learning}

MLIR has become the foundation of several modern compiler stacks targeting machine learning workloads. The principal aim is to create a unified system that can take models from multiple frameworks (PyTorch, TensorFlow, ONNX, etc.) and progressively lower them using MLIR, enabling optimization at different levels of abstraction. These systems differ in how they use MLIR, which dialects they optimize, and how optimization decisions are made.

\subsubsection{IREE}
IREE \cite{iree} is an end-to-end compiler and runtime system designed to lower machine learning models to a unified intermediate representation. It relies heavily on the \emph{Linalg} dialect as its core abstraction for compute-intensive operations. IREE’s compilation flow lowers high-level inputs (from TensorFlow, JAX, PyTorch, etc.) into \emph{Linalg} operations, where it performs hardware-agnostic transformations such as fusion and tiling. These transformations are typically applied via a predefined pass pipeline that utilizes static heuristics and target-specific configurations to determine tile sizes and vectorization factors. The code is eventually lowered to the HAL (Hardware Abstraction Layer) dialect for execution on diverse backends including CPUs, GPUs, and micro-controllers.

\subsubsection{ONNX-MLIR}
ONNX-MLIR \cite{jin2020compiling} is a compiler specifically designed to compile valid Open Neural Network Exchange (\emph{ONNX}) graphs into optimized binary code. It introduces two specific dialects: the \emph{ONNX} dialect, which maps one-to-one with the \emph{ONNX} standard specification, and the \emph{Krnl} dialect, which serves as a bridge for loop-level optimizations. Optimizations are applied at the \emph{Krnl} dialect level. The \emph{Krnl} dialect allows for the explicit representation of schedules enabling transformations such as tile, skew, and transpose. Optimization decisions, such as loop ordering and tile sizing, are generally driven by static analysis and lowering rules implemented within the compiler's conversion passes.

\subsubsection{XLA}
XLA \cite{openxla} is a domain-specific compiler originally built for TensorFlow but now supporting JAX and PyTorch. XLA implementations increasingly leverage MLIR infrastructure, particularly through the MHLO (MLIR HLO) and StableHLO dialects, which represent high-level linear algebra operations. XLA’s primary optimization strategy revolves around aggressive kernel fusion to reduce memory bandwidth usage, along with layout assignment and buffer analysis. The compiler selects optimizations using analytical cost models and greedy heuristics to decide which operations to fuse or how to layout memory.\\

While IREE, ONNX-MLIR, and XLA are robust production compilers, they primarily rely on static heuristics and predefined pass pipelines to select optimizations. For instance, IREE uses target configurations to guide tiling, and XLA uses a cost model for fusion. In contrast, \toolname{} replaces these fixed heuristics with a learned policy. By exposing the \emph{Linalg} dialect transformations (tiling, fusion, interchange) as a multi-discrete action space, \toolname{} allows a Reinforcement Learning agent to explore and discover optimization schedules automatically. Unlike ONNX-MLIR’s \emph{Krnl} dialect which requires explicit scheduling logic, or XLA’s greedy fusion, our system enables the agent to learn which sequence of transformations to apply and how, based on empirical feedback, potentially uncovering optimization sequences that static heuristics might miss.

\section{Overview of \toolname{}}

Our RL framework uses the actor-critic RL architecture and is defined by the following key components:
1) \emph{Action space (detailed in Sec. \ref{section:action_space}):} defines the set of code optimizations that the agent can apply to the program being optimized;
2) \emph{State representation (detailed in Sec. \ref{states}):} provides a representation of the program being optimized;
3) \emph{Reward model (Sec. \ref{reward}):} provides a function that scores the quality of actions, guiding the agent toward actions that produce faster code;
4) \emph{Actor-critic networks (Sec. \ref{policy}):} the actor selects the next code optimization to apply given the current program state, and selects its parameters, while the critic estimates the expected return and provides the learning signal for policy optimization during training (the critic is only used during training to train the policy).

We train our proposed RL agent on \emph{a dataset} that we describe in Sec. \ref{data}, using the \emph{PPO learning algorithm}.

In this project, we focus on optimizing operations expressed in the Linalg MLIR dialect. The Linalg MLIR dialect provides a structured IR for linear-algebra and deep learning operations (e.g., matmul, convolutions, and generic elementwise and reduction kernels), defined over tensors or buffers with explicit iteration spaces and affine indexing maps. We chose the Linalg dialect because of its robustness, maturity, and wide set of code transformations that it supports. Early in the project, we considered other dialects, such as the Affine and the SCF dialects, which operate at a lower level (loop nests), but we faced difficulties due to the limited number of transformations that are well supported in those dialects. In addition, the main domains that we want to optimize (deep learning and LQCD computations) can both be fully expressed in this dialect. Listing~\ref{lst:linalg-generic} shows an example of a matrix multiplication implemented using the Linalg generic operation.

Although in this work we focus on optimizing the Linalg dialect, the design of \toolname{} is dialect-independent, in principle. The environment, reward interface, state representation, and RL infrastructure are fully reusable across MLIR dialects. Extending the system to other dialects (e.g., Tosa or Affine) mainly requires adapting the action space and the feature-extraction module, rather than redesigning the full framework. More precisely, our framework relies on the following assumptions: 1) the IR representation is at the loop level; 2) computations are sequences of loop nests (for-loops) manipulating arrays; 3) the dialect exposes a set of high-level loop-nest transformations; 4) the dialect provides a mechanism to guarantee the correctness of transformations, either because transformations are correct by construction or because legality is checked after dependence analysis. 
A dialect that satisfies these assumptions would, in principle, be compatible with \toolname{}, although demonstrating the effectiveness of our framework in those cases is left for future work.

{
\shrink[1]
\SmallCodeSize
\begin{lstlisting}[caption={MLIR \texttt{linalg.generic} operation}, label={lst:linalg-generic}]
linalg.generic {
    indexing_maps = [
      affine_map<(d0,d1,d2) -> (d0,d2)>,
      affine_map<(d0,d1,d2) -> (d2,d1)>,
      affine_map<(d0,d1,d2) -> (d0,d1)>
    ],
    iterator_types = ["parallel", "parallel", "reduction"]
}
  ins(%A, %B: memref<256x1024xf32>, memref<1024x512xf32>)
  outs(%C: memref<256x512xf32>) {
  ^bb0(%a: f32, %b: f32, %c: f32):
    %d = arith.mulf %a, %b : f32
    %e = arith.addf %c, %d : f32
    linalg.yield %e : f32
}
\end{lstlisting}
}

Since code usually contains multiple operations, \toolname{} processes one operation at a time. Operations are traversed in reverse order (from consumers to producers), as the Linalg fusion transformation has limited ability to fuse a modified producer. Starting from the consumer thus preserves more fusion opportunities. When multiple producer operations exist for a given consumer operation, we select the last producer (the one that occurs in the code right before the consumer, textually) as the next to fuse with the current consumer being optimized.

\section{Reinforcement Learning Environment}

In this section, we detail the components of our proposed RL environment. Specifically, we discuss the action space of the environment, the state and observations, and the proposed reward functions. We will use the notation presented in Table~\ref{tab:constraints} throughout the rest of the paper.

\begin{table}[htbp]
\small
\centering
\begin{tabular}{ll}
\hline
\textbf{Symbol} & \textbf{Description} \\ \hline
$N$             & Number of loop levels in a loop nest\\
$M$             & Number of tile sizes \\
$D$             & Rank of matrix accesses \\
$L$             & Number of accessed matrices \\
$\tau$          & Length of a transformation sequence \\ \hline
\end{tabular}
\caption{Notation used in the paper.}
\label{tab:constraints}
\shrink[2]
\end{table}

\subsection{Action Space}
\label{section:action_space}

An action \( a_t \in A \), where \( A \) is the action space, allows the agent to transition from a state \( s_t \) to another state \( s_{t+1} \). In this project, an action is a code transformation (code optimization) that can be applied to a loop nest (an operation). We focus on the following transformations:

\begin{itemize}
    \item \textbf{Tiling:} This transformation allows the working set of data to fit better into the cache, thereby improving memory access patterns and overall performance.
    We use the notation \( T(t_1, t_2, \ldots, t_N) \) to specify that tile size \( t_i \) is used to tile the loop level \( i \).
    Note that a tile size of zero indicates no tiling.
    
    \item \textbf{Tiled Parallelization:} This transformation in the Linalg dialect is the combination of two transformations: tiling followed by the parallelization of the outermost loop. One can apply parallelization alone, without tiling, by selecting a tile size of 1 for all the loop levels. Parallel execution is achieved by generating \emph{scf.parallel} operations, which are lowered to the \emph{OpenMP} dialect (specifically \emph{omp.wsloop}) and subsequently to \emph{LLVM IR}, inserting the necessary runtime calls to the \emph{OpenMP} library to manage thread creation and synchronization.

    \item \textbf{Tiled Fusion:} This transformation is the combination of tiling followed by a fusion. It merges a producer loop, which generates data in the form of a tensor, with a consumer loop that accesses that data. This serves to reduce the need for storing intermediate results, which can improve data locality.

    In Linalg, a loop must be tiled first, before being fused. This is because fusion is implemented at the tile granularity of the consumer. One first tiles the consumer, which creates explicit outer loops over tiles. Only then can the producer be cloned and moved inside those outer tile loops so each tile computes its needed portion locally.

    \item \textbf{Interchange:} Interchange involves permuting the order of loop levels using a permutation denoted by \( I(a_1, a_2, \ldots, a_n) \), where \( a_i \) represents the new order (index) of loop \( i \). For example, if we have a loop nest with 3 loop levels, the interchange \( I(2, 0, 1) \) will move the innermost loop to become the outermost.

    \item \textbf{Vectorization:} vectorizes the innermost loop nest.

    \item \textbf{No Transformation:} A special action that allows the agent to stop optimizing the current operator being optimized (current consumer being optimized) and move to the next operator (one of the producers of this consumer operator). When there is no producer left to optimize, the agent stops.
\end{itemize}

To illustrate the effect of these optimizations on MLIR code, Listing~\ref{lst:matmul-optimized} shows the result of applying \emph{Tiled Parallelization} with tile sizes \([8, 8, 0]\), followed by \emph{Vectorization}, to the operation in Listing~\ref{lst:linalg-generic}.

Each action is defined by a transformation and an associated set of parameters (if required). The sizes of the action space for each transformation are as follows:  

\begin{itemize}
    \item \textbf{Tiled transformations (\textit{Tiling}, \textit{Tiled Parallelization}, \textit{Tiled Fusion}):} \(M^N\).  
    \item \textbf{Interchange:} \(N!\), representing the number of possible loop permutations.  
    \item \textbf{Vectorization and No Transformation:} 1, since no additional parameters are required.  
\end{itemize}

The total size of the action space \(|A|\) is:  {\CodeSize $|A| = 3 \cdot M^N + N! + 2$}

\subsubsection{Multi-Discrete Formulation}
\label{sec:multi-discrete}

The above action space is large, and learning a policy for such a large action space is not trivial.
To address this, we reformulate the above action space into a \textit{multi-discrete action space}, where an action is represented as the Cartesian product of multiple sub-action spaces (i.e., an action is the combination of multiple simpler sub-actions). As a result, the action space becomes a composition of smaller discrete distributions defined as follows:  

\begin{itemize}
    \item \textbf{Transformation Selection:} A categorical distribution over six possible transformation options (Tiling, Tiled Parallelization, etc.).
    
    \item \textbf{Tiling Sizes for each Tiled Transformation:} The goal is to select a tile size for each loop level; therefore, we define \(N\) categorical distributions (one per loop level), each over \(M\) candidate tile sizes.
    \item \textbf{Interchange:} We propose two formulations for interchange:  
    \begin{itemize}
        
        \item \textbf{Enumerated Candidates:} We enumerate a subset of loop interchange candidates by swapping two loop levels that are either adjacent or separated by one or two levels. This restricted enumeration aims to keep the action space tractable, substantially reducing the number of possibilities while still capturing useful transformations.
        \item \textbf{Level Pointers:} Inspired by the work of Vinyals et al. \cite{vinyals2015pointer}, the interchange action is decomposed into a sequence of sub-steps. At each step \(i \in [0, N)\), the agent selects a loop \(n \in [0, N)\) to be placed at position \(i\). This requires a probability distribution over \(N\) loops. This method manages to cover all possible permutations, without the need to enumerate all the possibilities.
    \end{itemize}
\end{itemize}

\subsubsection{Action Mask} 
Not all actions are valid at every time step. To address this, we provide the agent with an \textit{action mask} that filters out invalid actions. The mask is created based on the current state. An example of actions that need to be masked is the vectorization action when the innermost loop is too large (having more than 512 iterations). In such a case, the MLIR vectorization pass leads to the generation of large and inefficient code, as the vectorization pass in MLIR fully unrolls the inner loop. This causes excessive memory consumption and significantly slows down the RL training.

\subsection{States and Observations\label{states}}

\begin{figure*}[htbp]
    \centering
    \includegraphics[width=0.9\textwidth]{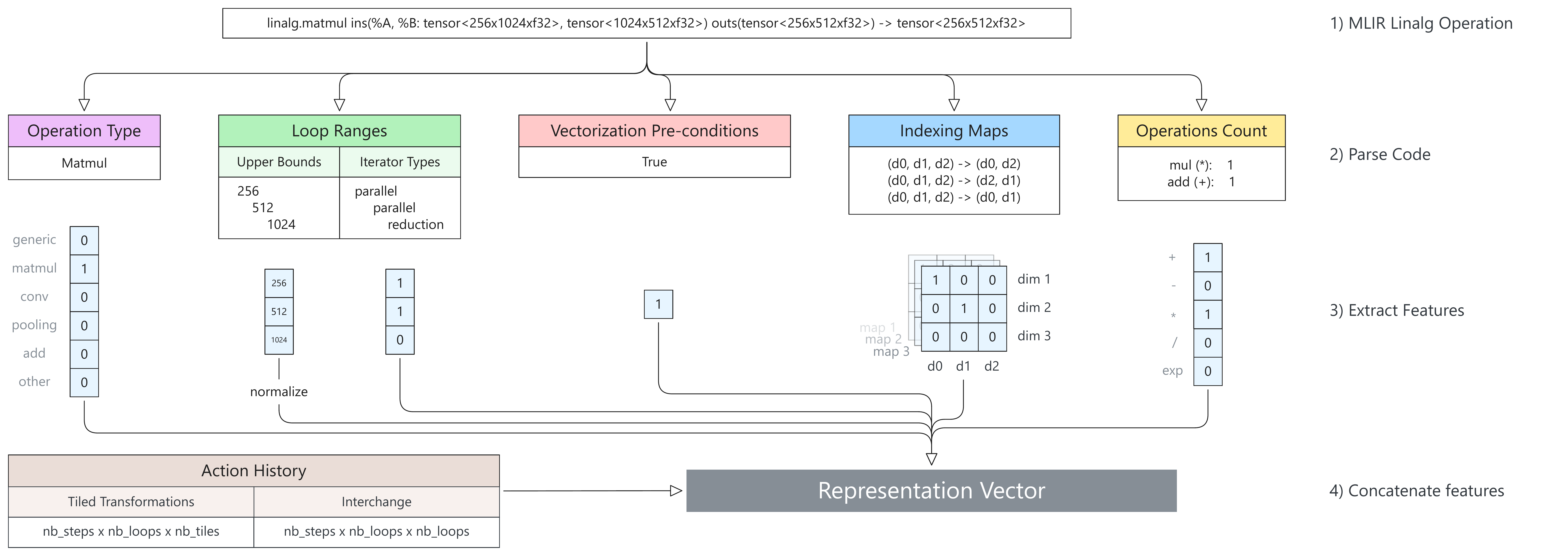}
    \caption{The pipeline of extracting the features from a Linalg operation and building the representation vector.}
    \label{fig:MLIR_feature_extraction}
    \shrink[1]
\end{figure*}

A linalg code is composed of a sequence of linear algebra operations and loop nests (implemented using the \emph{generic} linalg operator). Each operation and loop nest in the code is represented with a vector of features. Each vector is the result of concatenating a set of features (shown in Figure~\ref{fig:MLIR_feature_extraction}). We present these features in this section.

\begin{itemize}
    \item \textbf{Operation Type:} A one-hot vector encoding the Linalg operation type. The considered types include: matmul, conv\_2d, add, pooling, and generic, with an additional unknown category to account for any other operation type that was not seen during training and that the system may encounter. Generic operations in the linalg dialect are used to express general loop nests. Note that some common operations such as \emph{ReLU} don't exist in the Linalg dialect so they are explicitly coded using \texttt{linalg.generic} which means that their type here is also considered \emph{generic}.
    Using named operations, when available, helps the agent learn optimization patterns that are specific to particular loop-nest structures, which can lead to better optimization decisions than treating all operations as fully generic.

    \item \textbf{Loop Ranges:} We extract two key characteristics of each loop: the upper bound and the iterator type. The lower bound and step are omitted, as they are always fixed to 0 and 1, respectively, in Linalg operations. The iterator type  (reduction or parallel iterator) is essential for determining the legality of parallelization, depending on whether the iterator is used for a reduction (\textit{reduction}) or not (\textit{parallel}).

    \item \textbf{Vectorization Pre-conditions:} In MLIR, not all Linalg operations are eligible for vectorization, as specific conditions must be satisfied beforehand. If these conditions are not met, vectorization attempts will fail. To account for this, we include a boolean flag indicating the outcome of these pre-condition checks, thereby informing the agent whether vectorization is feasible for a given operation.

    \item \textbf{Indexing Maps:} Each Linalg operation is associated with indexing maps that specify how data is read or written in tensors. An indexing map defines the mapping between loop iterators (\(d_0, d_1, \ldots, d_N\)) and the indices of the accessed tensor, expressed as an affine function of the iterators for each dimension. For example, the map \texttt{(d0, d1, d2) -> (d0 + 1, 3 * d2)} indicates that the operation has three iterators (three loop levels), where the first tensor dimension is accessed at \(d_0 + 1\) and the second at \(3 \cdot d_2\). Inspired by work done by Baghdadi et al. \cite{baghdadi2021deep}, we represent each map as a polyhedral access matrix of size \(D \times N\), where each row corresponds to one dimension of the tensor, and each column corresponds to a loop iterator. The entries of the matrix capture the coefficients of each loop iterator.

    \begin{figure}[htbp]
        \centering
        \includegraphics[width=0.8\linewidth]{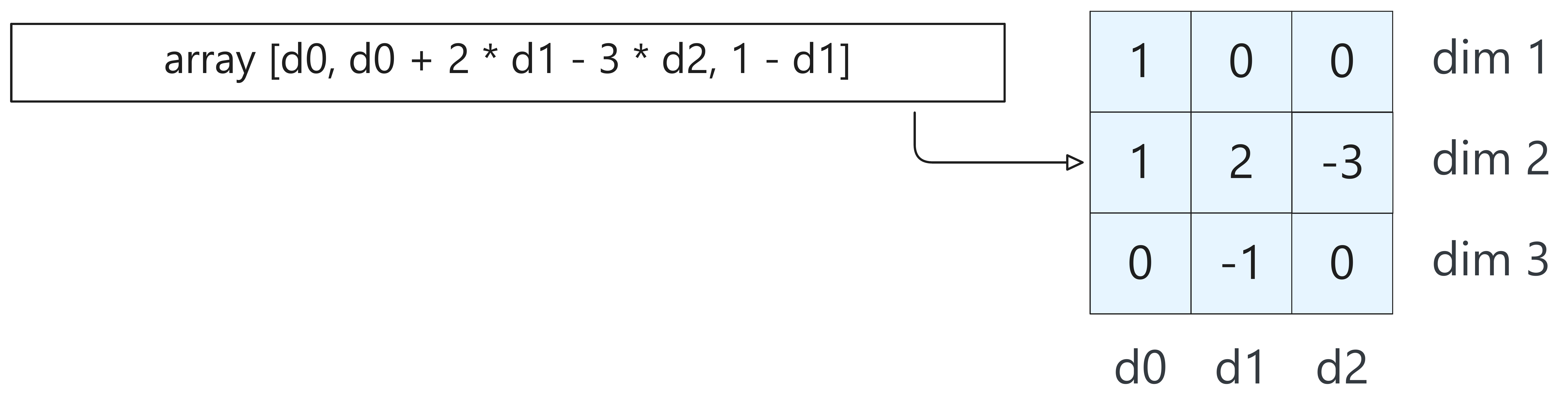}
        \caption{Example of an access matrix.}
        \label{fig:access_matrix}
        \shrink[1]
    \end{figure}

    \item \textbf{Operations Count:} We count the number of each arithmetic operation, including: addition (+), subtraction (-), multiplication (*), division (/), and exponential (exp). We store these numbers in a vector that represents operation counts.

    \item \textbf{Action History:} We record the sequence of applied transformations by storing their parameters (when they exist) in one-hot encoded matrices, indexed by time. At each time step, if a transformation is not applied, the corresponding entry is zero; otherwise, it stores the chosen parameters. More details on the exact encoding for each transformation type are provided in Appendix~\ref{appendix:action-history}.
    
\end{itemize}

\shrink[1]
\subsection{Reward\label{reward}}

An intuitive reward for the task of code optimization is the speedup of the optimized code (acceleration rate), which is the ratio between the old execution time to the new one. However, since the goal of reinforcement learning is to maximize cumulative rewards, we chose to use the logarithm of the speedup. This approach leverages the additive property of logarithms, making it more suitable for reward accumulation.

The reward is given at the end of an episode. We assign a reward of 0 after every step except for the last step, where we execute the optimized code and return the speedup.
    
We have also tried another reward structure where intermediate rewards are provided after each step, in addition to the above terminal reward, but the use of such intermediate steps does not improve the policy yet it makes the training slow (because we need to execute the optimized code after each step, to calculate the reward).

\shrink[1]
\section{Actor-Critic Network Architecture\label{policy}}

\begin{figure*}[htbp]
    \centering
    \includegraphics[width=0.85\textwidth]{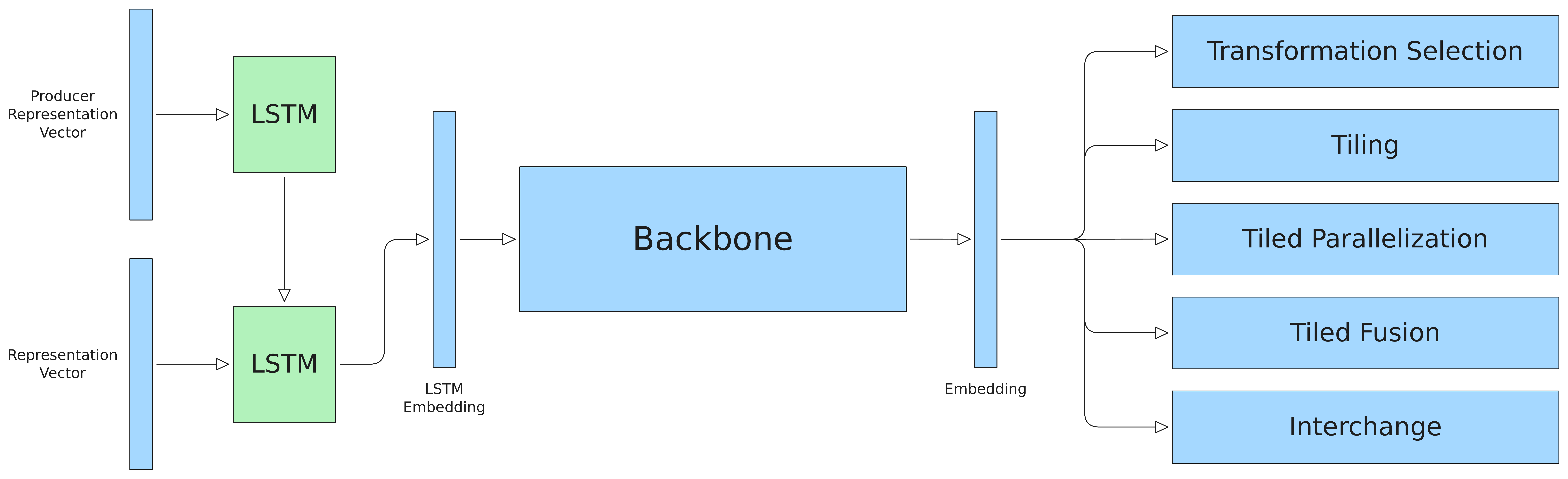}
    \caption{The RL agent’s policy network architecture consists of a backbone that processes the input representation vector into a feature vector that is then passed to the subnetworks to predict the transformation to apply and its parameters.}
    \label{fig:RL_architecture}
    \shrink[1.5]
\end{figure*}

\begin{figure*}[htbp]
    \centering
    \includegraphics[width=0.9\textwidth]{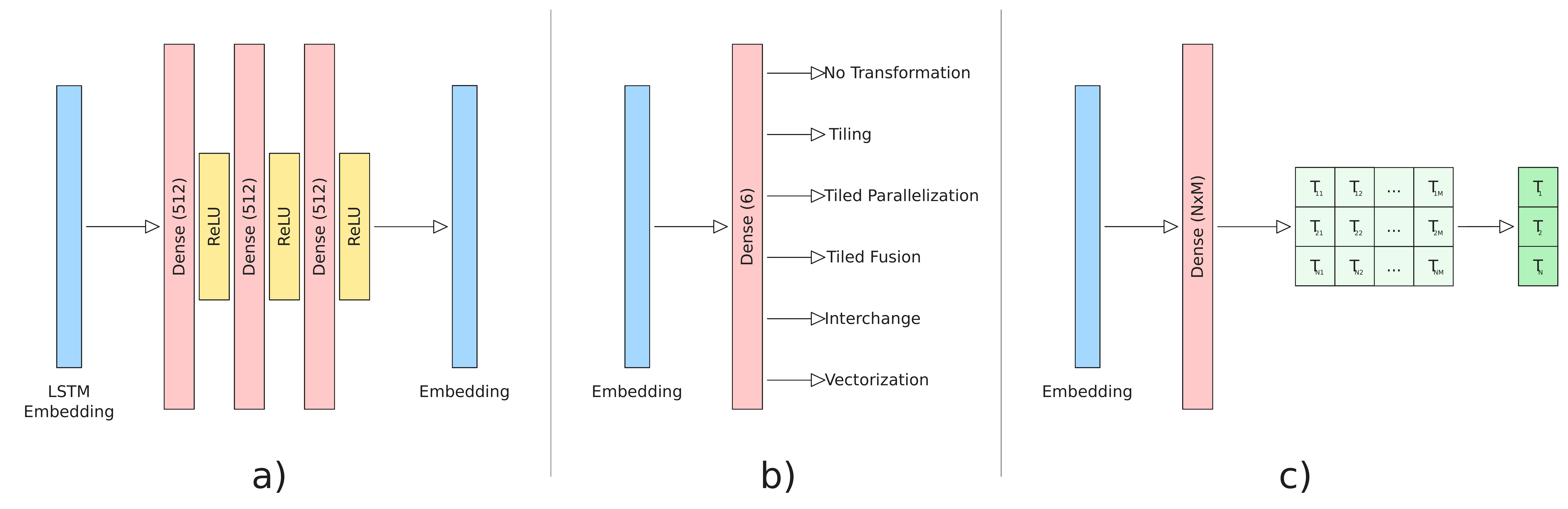}
    \caption{The detailed architecture of the networks used in the policy network: a) The backbone; b) The transformation selection network; c) Tiled transformations network. The interchange network varies in size depending on the interchange method, but it is always a dense layer outputting one distribution.}
    \label{fig:all_networks}
    \shrink[1.5]
\end{figure*}

A central component of Reinforcement Learning is the \textit{agent}, the entity responsible for selecting actions given input states. In this section, we describe our agent, designed for the previously introduced MLIR environment. We adopt an actor-critic architecture, which decomposes the agent into two sub-components: the \textit{actor}, responsible for learning a policy to select actions, and the \textit{critic}, responsible for evaluating how good a state is under the actor's policy.

The actor is the policy network, which learns a policy \(\pi\) that maps each state \(s \in S\) to a probability distribution over actions.
Formally, \(\pi(a|s)\) denotes the probability of taking action \(a \in A\) when in state \(s\).

The critic is the value network, which estimates the value function \(v_\pi(s)\), defined as the expected return when starting from state \(s\) and following policy \(\pi\).

\subsection{Policy Network}

Since the action space is partitioned into multiple subspaces, the policy network must sample a sub-action from the subspaces to construct the final action. We first sample the type of the transformation to apply (one of 6 transformation options), then we sample the parameters of the selected transformation. To implement this, we use a single policy network that has three components: a network to extract an embedding for the code being optimized, a backbone to create a rich embedding from the input embedding, and multiple heads to map the rich embedding to sub-actions (one head for selecting the transformation, and multiple other heads for selecting the parameters of transformations, where each head is specialized in selecting the parameters of a given transformation).

More precisely, the policy network is composed of the following three components:  

\begin{enumerate}
    \item \textbf{Producer-Consumer Embedding:} 
    The goal of this component is to take the code representation as input and produce an embedding for that code.
    Instead of creating an embedding for the whole code, which would require passing a large input to the model, making learning hard, we extract the representation of two Linalg operators (a consumer and its producer). The goal is to make the input size smaller, which would simplify learning. Passing the representation of two operators is sufficient because we optimize code operator by operator, starting from the last consumer in the code. While optimizing this single operator, we decide whether to fuse it with its producers. 
    Therefore, at a given time, we only need to pass the representation of two operators.
    To create an embedding that captures the relation between the producer and the consumer, we feed their representation vectors sequentially into an LSTM layer with 512 units. The final hidden state of the LSTM is used as the producer-consumer embedding.
    Although our preliminary tests showed that a dense network yields similar results, we selected an LSTM architecture because it naturally supports future extensions towards multi-producer fusion.

    \item \textbf{Backbone Network:} The LSTM embedding is passed through a backbone consisting of three fully connected layers with 512 neurons each, activated by ReLU \cite{activation} (see Figure~\ref{fig:all_networks}.a).  
    \item \textbf{Action Heads:} The embedding created by the backbone is fed into five different output heads:  
    \begin{itemize}
        \item \textbf{Transformation Selection:} A fully connected layer of size 6 (corresponding to the number of transformation options), followed by a Softmax activation \cite{activation}, which produces a probability distribution over the transformation options (see Figure~\ref{fig:all_networks}.b).  
        \item \textbf{Tiled Transformations:} Three heads, each consisting of a fully connected layer of size \(N \times M\), reshaped into a 2-D matrix. Each row of the matrix is independently activated with Softmax, producing one distribution per loop level, each over \(M\) candidates, to select a tile size for each loop (see Figure~\ref{fig:all_networks}.c).
        
        \item \textbf{Interchange:} A fully connected layer followed by Softmax. The size of this layer depends on the chosen interchange method: \(3N-6\) for enumerated candidates, or \(N\) for level pointers (see Section~\ref{sec:multi-discrete} for definition and Appendix~\ref{appendix:levels-pointers} for a detailed explanation).

    \end{itemize}
\end{enumerate}

\subsection{Value Network}

The value network also uses three components. The first two components are identical to the first two components of the policy network (the producer-consumer embedding and the backbone network). The backbone embedding is then passed through a fully connected layer with a single output neuron, which estimates the state-value function \(v_\pi(s)\).

\shrink[0.5]
\section{Dataset\label{data}}

Our goal is to train our agent to optimize code from two domains: deep learning and LQCD computations. Therefore, we created a dataset composed of code from these two domains. The total size of the dataset is 3959 training examples. The rest of the section describes how we created the dataset.

\shrink[0.5]
\subsection{Deep Learning Data}

We curated a new training dataset specifically for our task. Since our objective is to optimize full neural network code, we required a dataset composed of two types of computations: single deep learning operators and sequences of deep learning operations similar to those found in real neural network models. To construct this dataset, we adopted two approaches. First, we collected commonly used deep learning operators from open source models; Second, we randomly synthesized sequences of deep learning operations.
To keep the training computationally feasible, we limited the sequence lengths to \texttt{L}=5, which provides a balance between keeping the training time reasonable while still allowing the agent to learn how to handle multiple operations simultaneously. Note that the agent, while optimizing a sequence of operators, only represents (perceives) two operators at a time and does not represent (perceive) the whole sequence of operators.

\begin{table}[h!]
\CodeSize
\centering
\begin{tblr}{
  column{2} = {c},
  column{3} = {c},
  hline{1-2,7} = {-}{},
}
\textbf{Operation}    & \textbf{Training set} \\
Matrix multiplication & 187                   \\
2d convolution        & 278                   \\
Maxpooling            & 250                   \\
Matrix addition       & 271                   \\
ReLU                  & 149                   \\
Total                 & 1135                  
\end{tblr}
\caption{The distribution of each DNN operator in the single-operator training sets.}
\label{tab:data_collection}
\shrink[2]
\end{table}

To generate the dataset of single operators, we collected 121 state-of-the-art models from TensorFlow Hub and Hugging Face, spanning from vision models to transformers. We then collected the operations used in these models (e.g., conv, relu, matmul, etc.). From the collected operations, we selected the most frequently occurring ones. For each of
these operations, we generated many variants by varying the input sizes and shapes. Each operation variant became a training example in our training dataset. We collected a total of 1135 training examples. These operations span multiple types, including matrix multiplication, convolutions, max pooling, additions, and ReLU activation (more details about the composition of this dataset in Table \ref{tab:data_collection}).

To generate the random operator sequences, we generated a random sequence of length \texttt{L}=5. Each operation in the sequence takes as input the output of the previous operation. The operations are generated with random shapes and are randomly selected from the following set of operations: \texttt{add}, \texttt{matmul}, \texttt{relu}, \texttt{conv\_2d} and \texttt{pooling}, \texttt{sigmoid}, and \texttt{softmax\_2d}.

\shrink[0.5]
\subsection{LQCD Data}

We integrated \toolname{} as a backend for a DSL compiler developed for the LQCD domain (this compiler is still unpublished). LQCD is a method for studying the strong force by simulating quarks and gluons on a space-time grid. 
A common task in LQCD is to compute \emph{correlators}, which are mathematical objects used to study the properties of particles and their interactions. LQCD computations are usually a long sequence of loop nests that read and write to tensors. Loop nests are usually deep (reaching more than 12 loop levels), and some of the accesses to tensors are irregular. Many of the loop nests are parallel, and many of them contain reductions at the inner levels.

The LQCD compiler takes as input a DSL code that is embedded in Python. The LQCD compiler then lowers this code and generates Halide code as output. We have built a pass to translate Halide code to the MLIR Linalg dialect. We then use \toolname{} to optimize it.

The LQCD compiler has 7 tests, each test is a large LQCD code (thousands of lines) designed to test the ability of the compiler to compile patterns that are commonly found in LQCD codes. We extracted all the loop nests from these tests. For each of these loop nests, we generated many variants by varying the input sizes. Each loop nest variant became a training example in our training dataset. In total, we collected 691 training examples.

\section{Experiments and Results}

\subsection{Experimental Setup}

\subsubsection{Hardware and Software Configurations}

We perform our evaluations on a multicore dual-socket node, each socket is a 14-core Intel(R) Xeon(R) CPU E5-2680 v4 @ \SI{2.40}{GHz} with \SI{64}{GB} RAM total. We use \NbNodes{} nodes identical to this one for the training. The transformations are implemented using MLIR, built on LLVM release 19.

\subsubsection{Benchmarks}
We evaluate \toolname{} on three sets of benchmarks:
\begin{itemize}
    \item \emph{Deep Learning Operators}: we use a benchmark of common operations in neural networks, namely matrix multiplication, convolution, pooling, addition, and ReLU. We use input data sizes and shapes that we obtained from widely used models (e.g., ResNet). The data sizes and shapes used for the evaluation were not seen during training.
    \item \emph{Neural Network Models}: we also evaluated \toolname{} on a set of 3 neural network models: ResNet-18, VGG, and MobileNetV2. We use the PyTorch implementations of these models, and automatically generate the equivalent MLIR Linalg code using Torch-MLIR \cite{torch-mlir}. We provide more details about these models and their composition in Appendix~\ref{appendix:models}.
    \item \emph{LQCD Applications}: This benchmark suite includes three increasingly complex applications: 1) \emph{Dibaryon–Dibaryon:} A baryon is a three-quark bound state (e.g., proton or neutron). A dibaryon is two baryons bound together (six quarks total). Thus, this benchmark studies interactions between two dibaryons (i.e., two six-quark states);
    2) \emph{Dibaryon–Hexaquark:} comparing a two-baryon system with a six-quark exotic particle; and
    3) \emph{Hexaquark–Hexaquark:} the heaviest case with correlators between two six-quark states.
    These LQCD applications have a number of lines ranging from \emph{1000} to \emph{8000} lines of MLIR Linalg code. 
\end{itemize}

\subsubsection{Metrics}
We report the speedup of the optimized codes over the baselines. Our baselines are the MLIR implementations of the benchmarks compiled with the MLIR compiler without loop-level optimizations and with \emph{O3} applied automatically by LLVM (we use the same MLIR compiler pipeline that we use in \toolname{}, but we disable the selection of loop-level code optimizations done by \toolname{}). We run each code 5 times, and take the median of its \textit{execution times}. A speedup higher than 1 means that the optimized code is faster than the baseline unoptimized code.

\subsubsection{Comparison with state-of-the-art}

We compare \toolname{} to the following state-of-the-art compilers and frameworks. 

\begin{itemize}
    \item \emph{PyTorch}~\cite{pytorch}: this is the PyTorch framework version 2.8.0 used with the Intel MKL DNN library (v3.7.1). We evaluate PyTorch on the deep learning operators and deep learning models. We perform 10 warm-up executions, followed by 11 measurement runs. The execution time is recorded in each measurement run, and we report the median.
    \item \emph{PyTorch compiler}~\cite{pytorchjit}: a state-of-the-art industrial compiler for PyTorch. We evaluate the PyTorch compiler on the deep learning operators and deep learning models. The PyTorch compiler is invoked using \emph{torch.jit.script}, and we follow the same evaluation protocol as for PyTorch (warm-up followed by measurement runs).
    \item \emph{Halide RL}~\cite{pecenin2019optimization}: a reinforcement learning framework for the Halide compiler. We evaluate Halide RL on the deep learning operators. We do not evaluate it on full deep learning models or LQCD applications because its design makes supporting these cases difficult, and their system currently lacks automated tools to support these sources.
    \item \emph{Halide autoscheduler}~\cite{mullapudi2016automatically}: an autoscheduler for the Halide compiler, used as the default autoscheduler for optimizing the original LQCD applications. We evaluate it on the LQCD applications.
\end{itemize}

\subsubsection{RL Agent Training}

We train the agent for 10,000 steps (each step includes trajectory collection and 4 PPO update epochs). Training takes approximately 5 days and 7 hours on \NbNodes{} CPU nodes (node characteristics described above).

We set the maximum number of loop levels in a loop nest to \(12\), the number of tile sizes to \(8\) (including zero to represent no tiling), the maximum number of arrays accessed in each loop nest to \(14\), the maximum rank of array accesses to \(12\), and the maximum schedule length to \(5\).

The RL agent is trained using Proximal Policy Optimization (PPO) \cite{ppo} with a learning rate of 0.001 and a clip range of 0.2 to ensure stable policy updates. We set the discount factor to \(\gamma=1.0\) because rewards are delayed until the end of the trajectory, avoiding excessive discounting for long codes. The GAE parameter was set to \(\lambda=0.95\) to balance bias and variance in advantage estimation. Each trajectory corresponds to the full transformation sequence of all operations in a code sample. We collect trajectories from 64 code samples at a time, split the data into mini-batches of size 32, and perform 4 PPO update epochs. The value loss coefficient is set to 0.5 and the entropy coefficient to 0.01.

\subsection{Overhead of the Compilation Pass}

The overhead introduced by our compilation pass primarily comes from two sources: calls to the policy network and the application of MLIR transformations selected by the policy. The policy network is queried multiple times per program in order to generate the full optimization sequence. The average total inference time (CPU Only) is 0.028 \emph{s per code sample} (measured over single deep-learning operators and LQCD applications). Applying the selected transformations also incurs cost: for single deep-learning operators, MLIR takes on average 0.089 \emph{s per code sample} to apply the full sequence of transformations, while for LQCD applications the transformation time is 0.8 \emph{s per code sample}.

\subsection{Evaluation Results}

In this section, we present the results of our evaluation. We first present the results of our evaluation on the benchmark of deep learning operators, then our results on the deep learning models, and then on the LQCD applications.

\subsubsection{Evaluation on the Deep Learning Operators}

In this section, we show the evaluation of \toolname{} on the benchmarks of single deep learning operators. We compare the performance of \toolname{} to PyTorch~\cite{pytorch}, PyTorch compiler~\cite{pytorchjit} and Halide RL~\cite{pecenin2019optimization}. We present the speedups of code generated by these compilers over unoptimized MLIR code (baseline).

\begin{figure*}[htbp]
    \centering
    \includegraphics[width=\textwidth]{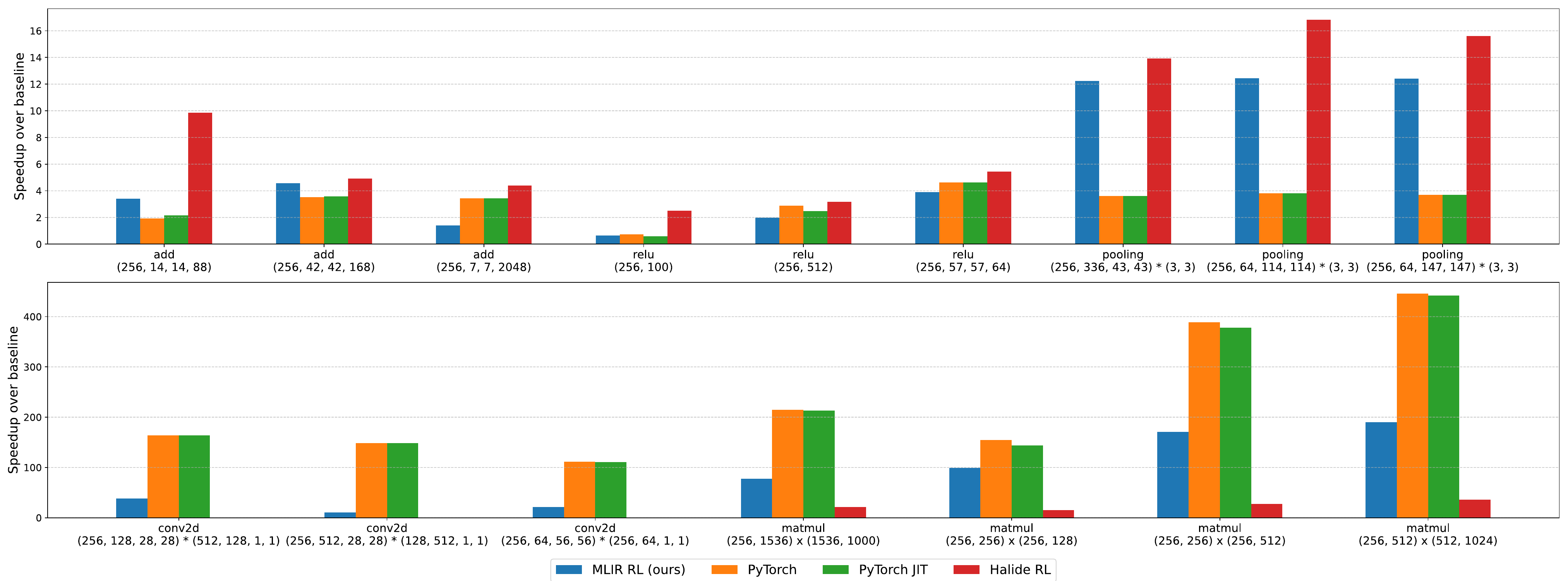}
    \caption{Speedups over MLIR baseline for each method across neural network operators, comparing our system (MLIR RL), Halide RL, PyTorch, and the PyTorch compiler.}
    \label{fig:dnnops}
    \shrink[2]
\end{figure*}

Figure~\ref{fig:dnnops} shows the results. On average, the relative performance varies significantly across operator types. For \emph{Add} and \emph{ReLU}, \toolname{} achieves competitive speedups compared to both PyTorch and the PyTorch compiler, showing that the agent can reliably learn effective schedules for elementwise operators whose performance is primarily bounded by memory throughput. For \emph{Maxpooling}, \toolname{} actually outperforms both PyTorch and the PyTorch compiler, achieving on average \(3.3\times\) higher speedups, mainly because \toolname{} automatically discovers effective tiling sizes that reduce memory traffic in these stencil-like operators for the target CPU.

For \emph{Conv2D} and \emph{Matmul}, however, \toolname{} generates code that is slower than PyTorch and PyTorch compiler (on average \(2.16\times\) slower on \emph{Matmul} and \(6.71\times\) slower on \emph{Conv2D}). This is because they leverage architecture-specialized kernels, register tiling, and aggressive vectorization implemented in libraries such as oneDNN. Although MLIR provides these capabilities, our current action space does not expose them yet, limiting the ability of the agent to optimize these compute-intensive programs.

Compared to Halide RL, \toolname{} achieves competitive results on \emph{Add} and \emph{ReLU}, but for \emph{Maxpooling}, Halide RL slightly outperforms \toolname{} on all sizes (\(1.25\times\) on average) due to the inability of our system to vectorize these operations. As for \emph{Matmul}, \toolname{} consistently achieves better speedups than Halide RL with an average of \(5.32\times\) indicating that \toolname{} manages to perform effective tilings and vectorization.

\subsubsection{Evaluation on the Deep Learning Models}

The following table presents the speedups over unoptimized MLIR code (baseline) achieved by our system (\toolname{}) in comparison to PyTorch and PyTorch compiler.

\begin{table}[ht]
\centering
\begin{tabular}{lccc}
\toprule
\textbf{NN Model}   & \textbf{\toolname{}}  & \textbf{PyTorch}  & \textbf{PyTorch compiler} \\
\midrule
ResNet-18           & 25.43                 & 374.77            & \textbf{411.26}           \\
MobileNetV2         & 6.93                  & 23.66             & \textbf{28.23}            \\
VGG                 & 54.64                 & 321.99            & \textbf{328.77}           \\
\bottomrule
\end{tabular}
\caption{Evaluating MLIR RL on different models.}
\label{tab:model-comparison}
\end{table}

For all three models, the highest speedup is achieved by PyTorch and PyTorch compiler. For ResNet-18, PyTorch compiler is \(16.17\times\) faster than \toolname{}. For MobileNetV2, PyTorch compiler outperforms \toolname{} by \(4.07\times\). For VGG, a \(6.02\times\) speedup is reached by PyTorch compiler.

These differences arise primarily from PyTorch's ability to achieve near-optimal performance in compute-intensive kernels (\emph{Matmul} and \emph{Conv2D}) which are the biggest bottleneck in these models. Additionally, our system still faces limitations vectorizing \emph{Conv2D} operations in these full models as it's not yet capable of converting them to \emph{Img2col + GEMM} which greatly limits its potential to achieve higher speedups.

\subsubsection{Evaluation on the LQCD Applications}

\begin{table}
\vspace{2mm}

\CodeSize
\centering
\begin{tblr}{
  column{2} = {c},
  column{3} = {c},
}
\hline
\textbf{Benchmark}                  & \textbf{\toolname{}}  & \textbf{Mullapudi} \\
\hline
hexaquark-hexaquark (S = 12)          & \textbf{13.25}    & 1.17 \\
dibaryon-dibaryon (S = 24)            & \textbf{7.57}     & 5.15 \\
dibaryon-hexaquark (S = 32)           & 2.15              & \textbf{4.68} \\
\hline
\end{tblr}
\caption{Comparison of speedups over MLIR baseline on LQCD benchmarks between \toolname{} (ours) and Halide autoscheduler (Mullapudi). \(S\) refers to the input size.}
\label{tblr:lqcd-results}
\shrink[1.5]
\end{table}

We compare the performance of \toolname{} with the Halide autoscheduler \cite{mullapudi2016automatically} on the benchmark of LQCD applications. Table \ref{tblr:lqcd-results} reports the speedups over unoptimized MLIR code (baseline).

The table shows that \toolname{} achieves competitive performance across the LQCD applications. In particular, for the \emph{hexaquark-hexaquark} and \emph{dibaryon-dibaryon} applications, \toolname{} outperforms Halide’s autoscheduler by up to $11\times$. These gains stem from \toolname{}’s ability to apply loop tiling to reduce memory traffic in the deep nests of the LQCD correlator codes. It also selects loop orders that expose unit‑stride vectorization and profitable outer‑loop parallelism.

\subsection{Ablation Study}

In this section, we present an ablation study composed of three experiments: 1) we compare our two proposed interchange methods \emph{Enumerated Candidates} and \emph{Level Pointers} (presented in Sec~\ref{sec:multi-discrete}); 2) we compare the use of a multi-discrete action space compared to a flat action space; 3) we compare the use of immediate reward compared to final reward.

For the first experiment, we train two agents, each with one of the proposed methods for loop interchange, and evaluate them on our benchmark suite. We observe that the agent trained with \emph{Level Pointers} achieves a higher average speedup of $18.7\times$ over MLIR unoptimized code, compared to $14.5\times$ for the \emph{Enumerated Candidates}. This highlights the efficiency of \emph{Level Pointers} in navigating the search space of interchanges.

\begin{figure}[htb]
    \shrink[1]
    \centering
    \includegraphics[width=0.53\linewidth]{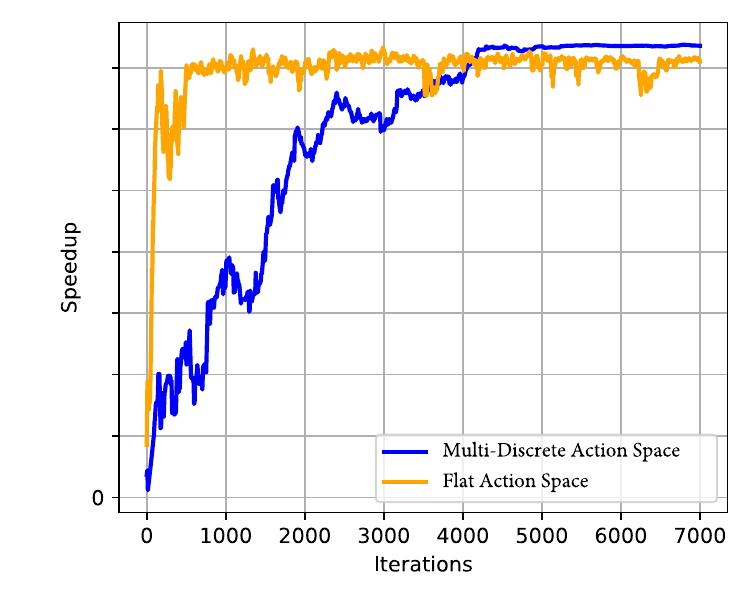}
    \caption{Comparison between the speedups achieved by training with a Flat Action Space and a Multi-Discrete Action Space.}
    \label{fig:Unrolled_vs_Hierarchical}
    \shrink[1.5]
\end{figure}

\begin{figure}[!ht]
    \centering
    \includegraphics[width=1.\linewidth]{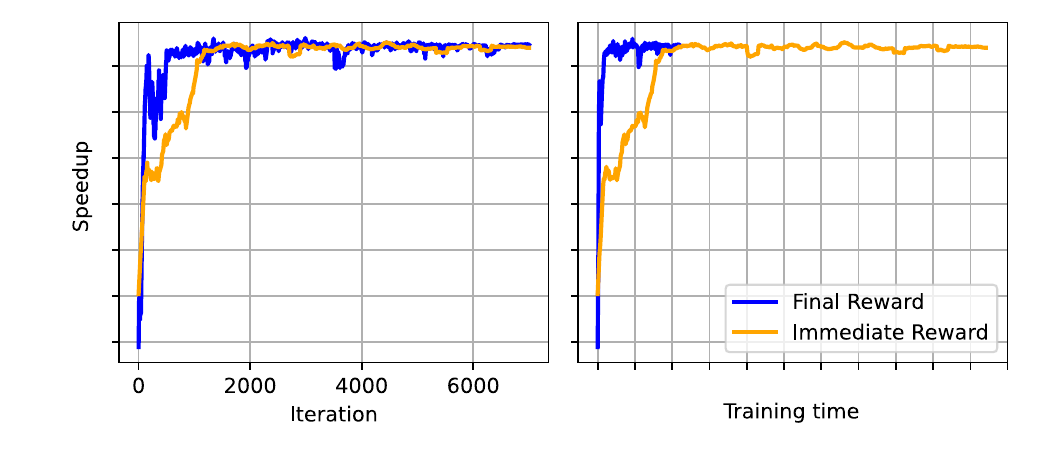}
    \caption{Comparison of "Immediate Reward" and "Final Reward" methods: the right plot shows the achieved speedup over training iterations; the left plot shows the achieved speedup over training time in hours.}
    \label{fig:Final_vs_Immediate_time}
    \shrink[2]
\end{figure}

For our second experiment, we compared two identical RL agents but using different action space configurations. The first model uses a simple, flat action space, which presents a fixed set of transformation combinations, where a transformation and its parameters represent a single action (we call it Flat Action Space); the flat action space therefore has a high number of actions. The second model uses our proposed Multi-Discrete Action Space, where  the agent first selects a transformation and then picks its parameters.

Our results are presented in Figure~\ref{fig:Unrolled_vs_Hierarchical}. The figure illustrates that while the Multi-Discrete Action Space model converges more slowly, it is capable of exploring a wider and more diverse range of actions, leading to a higher average speedup at the end. This might be explained by the fact that learning in the first case is simpler, since the agent has fewer possible actions in each step. The results suggest that the Multi-Discrete Action Space provides a more robust framework for discovering better optimizations, albeit at the cost of longer training times.

For the third experiment, we train two agents: one with an \emph{Immediate Reward} and the other with a \emph{Final Reward}, and evaluate them on the benchmarks. The results, shown in Figure~\ref{fig:Final_vs_Immediate_time}, indicate that both reward functions achieve comparable performance in terms of average speedup. However, given the substantial execution overhead introduced by the \emph{Immediate Reward} approach (as discussed in Section~\ref{reward}), we prefer to use \emph{Final Rewards}.

\shrink[0.5]
\section{Challenges and Future Work}

Developing MLIR RL presented two main challenges. First, designing an effective action space was time-consuming and required substantial domain expertise, amounting to roughly four person-years of effort. Second, training the RL agent is computationally expensive, and the need to retrain the agent many times during development made this a major bottleneck. Future work that reduces the cost of action-space design or accelerates training would be highly beneficial.

\shrink[0.5]
\section{Conclusion}
In this paper, we present \toolname{}, a Reinforcement Learning environment for automatic code optimization in the MLIR compiler. We propose a multi-discrete formulation of the action space where the action space is the Cartesian product of simpler action subspaces. We also propose a new method, called level pointers, to reduce the size of the action space related to the loop interchange transformation. These two methods enable more efficient and effective learning of the policy. Experimental results demonstrate that our proposed environment enables effective optimization of MLIR code, opening the door for more research on using RL for automatic code optimization.

\shrink[0.5]
\section{Data Availability}
The official code for MLIR RL is released in the following repository: \url{https://github.com/Modern-Compilers-Lab/MLIR-RL}.
The artifact evaluation archive is available in: \href{https://doi.org/10.5281/zenodo.17660414}{doi.org/10.5281/zenodo.17660414}.

\shrink[0.5]
\section{Acknowledgment}

This research has been partly supported by the Center for Artificial Intelligence and Robotics (CAIR) at New York University Abu Dhabi, funded by Tamkeen under the NYUAD Research Institute Award CG010. The research was carried out on the High-Performance Computing resources at New York University Abu Dhabi.

\section*{Artifact Appendix}

\subsection{Abstract}

This artifact contains \emph{\toolname{}}, a deep reinforcement learning (RL) system that optimizes loop nests in MLIR. It includes the RL environment, pre-trained models, MLIR benchmarks, and scripts required to reproduce all results for \toolname{} (Figure~\ref{fig:dnnops} and Tables~\ref{tab:model-comparison}–\ref{tblr:lqcd-results}) and the PyTorch / PyTorch compiler results in Figure~\ref{fig:dnnops} and Table~\ref{tab:model-comparison}.  

The artifact targets loop nests in the MLIR \texttt{linalg} dialect and uses a PPO-based actor–critic agent with a multi-discrete action space. It requires an LLVM/MLIR 19.x build (with Python bindings) and a Python 3.11 environment managed by Poetry. We provide a Docker image so experiments can be run on any Linux machine with Docker; the original experiments were run on an exclusive HPC node (Intel Xeon E5-2680 v4 @ 2.40GHz, 28 cores, 64\,GB RAM). Preparing the environment takes about 1 hour and reproducing all evaluation results takes less than 1 hour.

%%%%%%%%%%%%%%%%%%%%%%%%%%%%%%%%%%%%%%%%%%%%%%%%%%%%%%%%%%%%%%%%%%%%%
\subsection{Artifact check-list (meta-information)}

{\small
\begin{itemize}
  \item {\bf Compilation: } LLVM/MLIR 19.x, built from source (Release, MLIR+Python bindings, X86 target); Clang/LLD; CMake+Ninja.
  \item {\bf Transformations: } Loop tiling, tiled parallelization, tiled fusion, interchange, vectorization.
  \item {\bf Binary: } Custom C++ MLIR tools (\texttt{AstDumper}, \texttt{PreVec}); standard MLIR tools from LLVM.
  \item {\bf Model: } Pre-trained PPO policy network (loaded via Python).
  \item {\bf Data set: } MLIR code files (\(\approx 700\) MB) + JSON baseline timings; downloaded from \url{https://nyu.box.com/shared/static/5y1ilrccu3443dhcr854dt23uv0fysym.zip}.
  \item {\bf Run-time environment: } Linux; Python 3.11 + Poetry; optionally Docker (recommended).
  \item {\bf Hardware: } CPU-only; evaluated on Intel Xeon E5-2680 v4 @ 2.40GHz, 28 cores, 64\,GB RAM, exclusive node; similar multi-core x86-64 recommended.
  \item {\bf Run-time state: } Exclusive node, no competing jobs; stable CPU frequency for reliable timings.
  \item {\bf Execution: } Command-line scripts; \(\approx 1\) hour for all evaluation runs.
  \item {\bf Metrics: } Execution time; speedup over MLIR baseline.
  \item {\bf Output: } JSON files with per-benchmark speedups for \toolname{}, PyTorch, and PyTorch compiler; Regenerated Figure~\ref{fig:dnnops} and Table~\ref{tab:model-comparison}.
  \item {\bf Experiments: } Follow \texttt{README}, shell scripts in \texttt{scripts/}, config files in \texttt{config/}; no manual editing required.
  \item {\bf How much time is needed to prepare workflow (approximately)?: } \(\approx 1\) hour.
  \item {\bf How much time is needed to complete experiments (approximately)?: } \(\approx 1\) hour.
  \item {\bf Publicly available?: } Yes (Git repository + dataset link).
  \item {\bf Code licenses (if publicly available)?: } MIT Licence.
  \item {\bf Workflow framework used?: } Shell scripts + config files; optional Docker.
  \item {\bf Archived (provide DOI)?: } \href{https://doi.org/10.5281/zenodo.17660414}{doi.org/10.5281/zenodo.17660414}.
\end{itemize}
}

%%%%%%%%%%%%%%%%%%%%%%%%%%%%%%%%%%%%%%%%%%%%%%%%%%%%%%%%%%%%%%%%%%%%%
\subsection{Description}

\subsubsection{How delivered}

The artifact is delivered as a public Git repository containing:
\begin{itemize}
  \item \toolname{} source code (Python + C++ tools) and configuration files under \texttt{config/}.
  \item A \texttt{Dockerfile} to build a self-contained image.
  \item Shell scripts in \texttt{scripts/} for training, evaluation, and comparison.
  \item Instructions and a link to the benchmark archive (MLIR files + JSON baseline timings).
\end{itemize}

Users clone the repository, optionally build and run the Docker image, download the dataset into \texttt{data/}, and follow the \texttt{README}.

\subsubsection{Hardware dependencies}

\begin{itemize}
  \item Target: x86-64 CPU.
  \item Tested: Intel Xeon E5-2680 v4 @ 2.40GHz, 28 cores, 64\,GB RAM, exclusive node.
  \item Recommended: \(\geq 16\) physical cores and \(\geq 64\) GB RAM for stable timings.
  \item No GPU or accelerator required.
\end{itemize}

\subsubsection{Software dependencies}

\begin{itemize}
  \item Linux OS.
  \item Either:
  \begin{itemize}
    \item Docker (recommended), using the provided \texttt{Dockerfile}, or
    \item Conda/Miniconda with Python 3.11, CMake, Ninja, Clang/LLVM, LLD, Poetry (versions as in \texttt{README}).
  \end{itemize}
  \item LLVM/MLIR 19.x built from source with MLIR and Python bindings enabled.
  \item Python dependencies installed via \texttt{poetry install}.
\end{itemize}

\subsubsection{Data sets}

\begin{itemize}
  \item MLIR benchmarks (DL operators, DL models, LQCD kernels) as \texttt{.mlir} files.
  \item JSON files with baseline execution times and benchmark lists for training/evaluation.
  \item Distributed as a single archive (\(\approx 700\) MB) to be placed and unpacked in \texttt{data/}.
\end{itemize}

%%%%%%%%%%%%%%%%%%%%%%%%%%%%%%%%%%%%%%%%%%%%%%%%%%%%%%%%%%%%%%%%%%%%%
\subsection{Installation}

We recommend the Docker-based installation; the \texttt{README} gives full commands.

\paragraph{Docker (recommended).}
\begin{enumerate}
  \item Clone the repository and \texttt{cd} into it.
  \item \texttt{docker build -t mlir-rl-artifact .}
  \item \texttt{docker run -it mlir-rl-artifact}
  \item Inside the container, download and unzip the benchmark archive into \texttt{data/}.
\end{enumerate}

\paragraph{Without Docker (sketch).}
\begin{enumerate}
  \item Create and activate a Conda environment with Python 3.11 and install the listed development packages.
  \item Clone and build LLVM/MLIR (release/19.x) with MLIR+Python bindings.
  \item Set environment variables (\texttt{PATH}, \texttt{PYTHONPATH}, \texttt{LLVM\_BUILD\_PATH}, \texttt{MLIR\_SHARED\_LIBS}) as in the \texttt{README}.
  \item Build custom tools in \texttt{tools/} and create a \texttt{.env} file with their paths.
  \item Run \texttt{poetry install} in the repository root.
  \item Download and unzip the benchmark archive into \texttt{data/}.
\end{enumerate}

%%%%%%%%%%%%%%%%%%%%%%%%%%%%%%%%%%%%%%%%%%%%%%%%%%%%%%%%%%%%%%%%%%%%%
\subsection{Experiment workflow}

All experiments are script-driven:

\begin{itemize}
  \item \textbf{Training (optional)}: \texttt{./scripts/train.sh}
        Trains the RL agent from scratch (long, not required for AE).
  \item \textbf{Evaluation (\toolname{})}: \texttt{./scripts/evaluate.sh}
        Evaluates a pre-trained model on the evaluation benchmarks and writes logs under \texttt{results/}.
  \item \textbf{Paper results}: \texttt{./scripts/paper.sh}
        \begin{itemize}
            \item Runs MLIR codes with no transformations to update the pre-recorded MLIR baseline.
            \item Runs \toolname{}, PyTorch, and the PyTorch compiler on the same benchmarks mentioned in the paper.
            \item Outputs speedups to JSON files in \texttt{paper/results/} (\texttt{mlir\_rl.json} for all \toolname{} results, \texttt{torch\_eager.json} for PyTorch, and \texttt{torch\_jit.json} for PyTorch Compiler); and Figure~\ref{fig:dnnops} and Table~\ref{tab:model-comparison} in \texttt{paper/figures/}.
        \end{itemize}
\end{itemize}

To reproduce the paper, evaluators:
\begin{enumerate}
  \item Run \texttt{paper.sh}.
  \item Observe the speedups in \texttt{paper/results/} and figures in \texttt{paper/figures/}.
\end{enumerate}

%%%%%%%%%%%%%%%%%%%%%%%%%%%%%%%%%%%%%%%%%%%%%%%%%%%%%%%%%%%%%%%%%%%%%
\subsection{Evaluation and expected result}

The artifact aims to reproduce:
\begin{itemize}
  \item \textbf{\toolname{} vs PyTorch/PyTorch compiler on operators} (Figure~\ref{fig:dnnops}):
        Average speedups over the MLIR baseline for key operators (e.g., Matmul, Conv2D, Pooling). The values for Halide RL are included directly (in order to recreate the same figure), as it's very complex and slow to reproduce results for Halide RL as well.
  \item \textbf{\toolname{} vs PyTorch/PyTorch compiler on models} (Table~\ref{tab:model-comparison}):
        Speedups for ResNet-18, MobileNetV2, and VGG.
  \item \textbf{\toolname{} on LQCD applications} (Table~\ref{tblr:lqcd-results}):
        Speedups for Dibaryon–Dibaryon, Dibaryon–Hexaquark, and Hexaquark–Hexaquark. The table isn't recreated here as the artifact doesn't reproduce Halide Autoscheduler results, due to the complexity of its setup.
\end{itemize}

Due to timing noise, we expect per-kernel execution times and speedups to vary within about \(\pm 5\%\) of the published numbers on similar hardware. For PyTorch and the PyTorch compiler, slightly larger deviations may occur: their performance depends on the exact PyTorch version, backend (oneDNN/MKL/OpenBLAS), and threading runtime, which can change kernel selection and fusion decisions.

%%%%%%%%%%%%%%%%%%%%%%%%%%%%%%%%%%%%%%%%%%%%%%%%%%%%%%%%%%%%%%%%%%%%%
\subsection{Experiment customization}

The system is configurable via JSON files in \texttt{config/}. Users can:
\begin{itemize}
  \item Change model and PPO hyperparameters (iterations, learning rate, batch sizes).
  \item Modify the action space (e.g., allowed transformations, maximum number of loops).
  \item Select different MLIR benchmarks.
  \item Adjust environment variables (e.g., \texttt{OMP\_NUM\_THREADS}) and hardware settings.
\end{itemize}

These knobs enable ablation studies and evaluation on additional benchmarks beyond those in the paper.

%%%%%%%%%%%%%%%%%%%%%%%%%%%%%%%%%%%%%%%%%%%%%%%%%%%%%%%%%%%%%%%%%%%%%
\subsection{Notes}

\begin{itemize}
  \item For AE, we recommend using the provided pre-trained models and focusing on reproducing evaluation results rather than retraining.
  \item Exclusive access to the node and stable CPU frequency significantly improve timing stability.
  \item The top-level \texttt{README} provides concrete command examples and troubleshooting tips.
\end{itemize}

\bibliographystyle{IEEEtran}
\bibliography{citations}

\appendix
\subsection{Action History Encoding}
\label{appendix:action-history}
For each transformation type, we maintain a separate one-hot encoded matrix, all sharing the same first dimension \(\tau\), the maximum sequence length. 
At time step \(t \in [0,\tau)\), if a transformation is not applied, the entire slice \([t,:,:]\) is set to zero; otherwise, the slice encodes the chosen parameters. 
\begin{itemize}
    \item \textit{Tiling:} stored in a matrix of shape \(\tau \times N \times M\), where \(N\) is the number of loops and \(M\) the possible tile sizes. The slice \([t,:,:]\) records the selected tile size \(m \in [0,M)\) for each loop \(n \in [0,N)\), or is all zeros if no tiling is applied.
    \item \textit{Interchange:} stored in a matrix of shape \(\tau \times N \times N\). The slice \([t,:,:]\) specifies which loop \(n \in [0,N)\) is placed in each position \(i \in [0,N)\), or is zero if interchange is not applied.
    \item \textit{Terminal actions (No Transformation, Vectorization):} no history is recorded, since applying them ends the optimization of the current operation and moves the agent to the next one. In case of vectorization, it's considered terminal since a vectorized Linalg operation gets completely replaced by vector operations thus disabling any further Linalg transformations.
\end{itemize}

\subsection{Level Pointers Detailed Explanation}
\label{appendix:levels-pointers}
\begin{figure}[h]
    \centering
    \includegraphics[width=\linewidth]{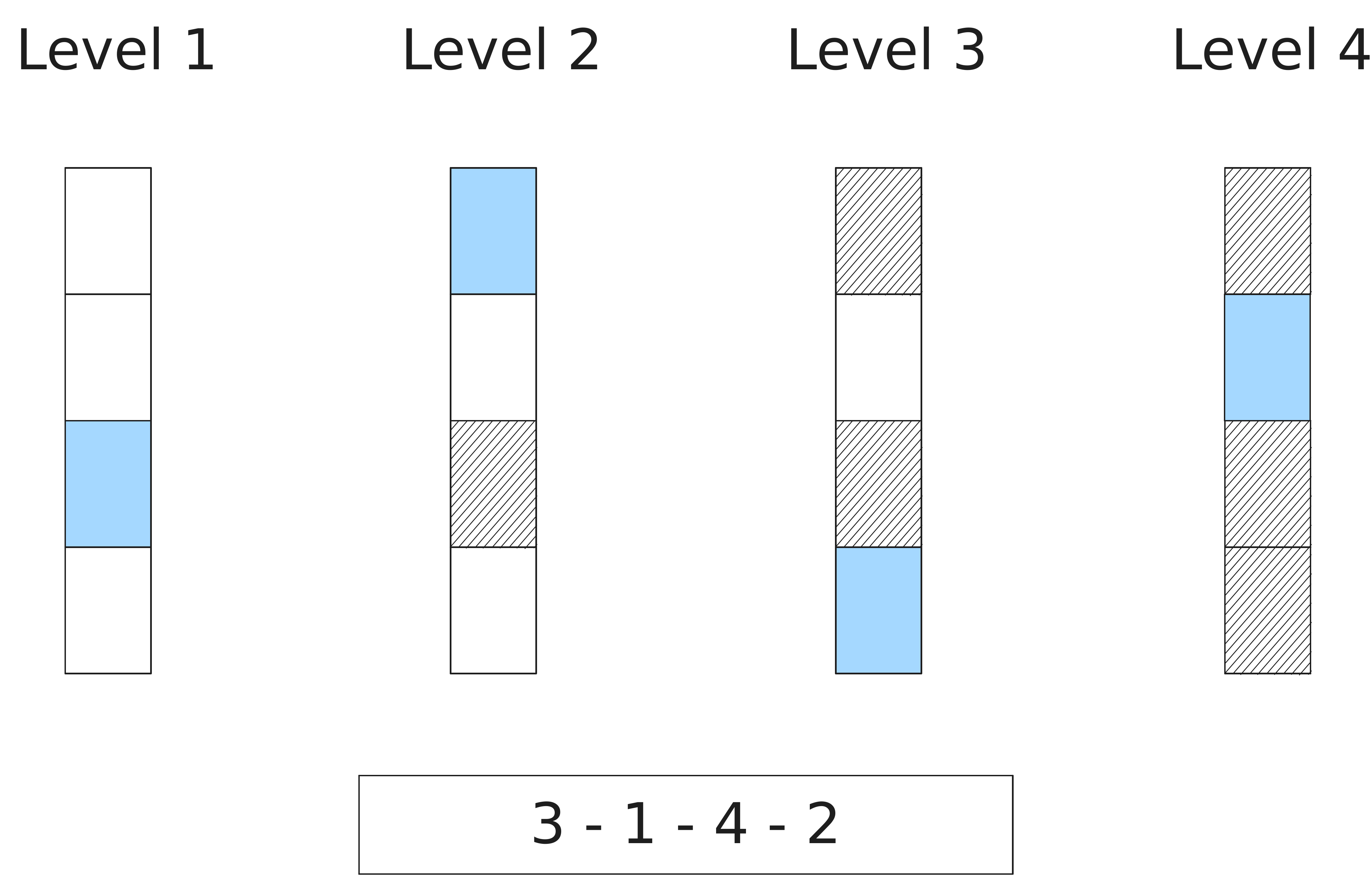}
    \caption{Illustration of the inference of interchange permutation using level pointers. The figure shows the output of the interchange head during the different sub-steps of the interchange action for a loop nest of size 4.}
    \label{fig:levels_pointers}
\end{figure}

For level pointers, completing an interchange action requires multiple inferences of the policy network. At the first step (when the agent decides to perform interchange), it selects a loop from the \(N\) available loops to be placed at position 0. At each subsequent time-step \(i \in [1, N)\), the agent is forced via the action mask to continue the interchange until completion. At each of these steps, it selects one of the remaining loops (with already chosen loops masked out) to be placed at position \(i\).
After \(N\) steps, this process produces a full permutation of the \(N\) loops (as illustrated in Figure~\ref{fig:levels_pointers}), which becomes the parameter of the interchange action. At this point, the agent is free to select any other transformation. To help the agent distinguish between intermediate sub-steps, partially selected loops are iteratively added to the interchange action history \ref{appendix:action-history}, giving the agent information about what has already been chosen and the current stage of the permutation.

\subsection{More Details about the Deep Learning Models\label{appendix:models}}

\begin{table}[hb]
\setlength{\tabcolsep}{3pt}
\centering
\begin{tabular}{lrrrrrr}
\toprule
\textbf{Model} & \textbf{Total} & \texttt{conv2d} & \texttt{pool} & \texttt{matmul} & \texttt{generic} & \texttt{unknown} \\
                & \textbf{Ops} &  &  &  &  &  \\
\midrule
MobileNetV2 & 524 & 35 & 1 & 1 & 448 & 39 \\
ResNet      & 510 & 53 & 2 & 1 & 438 & 16 \\
VGG         & 65  & 13 & 6 & 3 & 19  & 24 \\
\bottomrule
\end{tabular}
\caption{Operation composition of the benchmarked neural network models.}
\label{tab:model_composition}
\end{table}

We use three widely used neural network architectures for our evaluation.

\begin{itemize}
    \item \textbf{VGG (Visual Geometry Group):} VGG is a deep convolutional network known for its simplicity and uniform architecture. It primarily consists of stacked convolutional layers with small $3\times3$ kernels followed by max-pooling and fully connected layers. Despite its relatively high parameter count, VGG serves as a strong baseline for image classification tasks and is useful for evaluating performance on dense and repetitive patterns.

    \item \textbf{ResNet (Residual Network):} ResNet introduces residual connections that allow gradients to flow more effectively through very deep networks. This innovation enables the construction of extremely deep models without suffering from vanishing gradients. Its architecture contains residual blocks, each of which includes skip connections that bypass one or more layers. ResNet is widely used in both academic and industrial applications due to its robust performance and efficient training behavior.

    \item \textbf{MobileNet:} MobileNet is a lightweight neural network architecture optimized for mobile and embedded devices. It achieves efficiency by using depthwise separable convolutions, which reduce both the computational cost and the number of parameters. MobileNet is particularly suitable for performance benchmarking in low-resource environments, and it offers a useful contrast to the heavier VGG and ResNet models.
\end{itemize}

The composition of each model in terms of operation types is detailed in table ~\ref{tab:model_composition}.

\subsection{Linalg Operation After Optimization}

{\SmallCodeSize
\begin{lstlisting}[language=MLIR,caption={Optimized MLIR matmul after tiling and vectorization.},label={lst:matmul-optimized}]
#map = affine_map<(d0) -> (d0 * 8)>
#map1 = affine_map<(d0, d1) -> (d0, 0, d1)>
#map2 = affine_map<(d0, d1) -> (0, d1, d0)>
module {
  func.func private @nanoTime() -> i64 attributes {llvm.emit_c_interface}
  func.func @main(%arg0: tensor<256x512xf64>,
                  %arg1: tensor<512x1024xf64>,
                  %arg2: tensor<256x1024xf64>)
      -> (tensor<256x1024xf64>, i64)
      attributes {llvm.emit_c_interface} {
    %0 = call @nanoTime() : () -> i64
    %1 = scf.forall (%i, %j) in (32, 128)
         shared_outs(%acc = %arg2)
         -> (tensor<256x1024xf64>) {
      %i0 = affine.apply #map(%i)
      %j0 = affine.apply #map(%j)

      %A = tensor.extract_slice %arg0[%i0, 0]
           [8, 512] [1, 1]
           : tensor<256x512xf64> to tensor<8x512xf64>
      %B = tensor.extract_slice %arg1[0, %j0]
           [512, 8] [1, 1]
           : tensor<512x1024xf64> to tensor<512x8xf64>
      %C = tensor.extract_slice %acc[%i0, %j0]
           [8, 8] [1, 1]
           : tensor<256x1024xf64> to tensor<8x8xf64>

      %zero = arith.constant 0.0 : f64
      %vA = vector.transfer_read %A[0, 0], %zero
            {permutation_map = #map1}
            : tensor<8x512xf64> to vector<8x8x512xf64>
      %vB = vector.transfer_read %B[0, 0], %zero
            {permutation_map = #map2}
            : tensor<512x8xf64> to vector<8x8x512xf64>
      %vC = vector.transfer_read %C[0, 0], %zero
            : tensor<8x8xf64> to vector<8x8xf64>

      %mul = arith.mulf %vA, %vB : vector<8x8x512xf64>
      %red = vector.multi_reduction <add>, %mul, %vC [2]
             : vector<8x8x512xf64> to vector<8x8xf64>

      %C2 = vector.transfer_write %red, %C[0, 0]
             : vector<8x8xf64> to tensor<8x8xf64>

      scf.forall.in_parallel {
        tensor.parallel_insert_slice %C2
            into %acc[%i0, %j0] [8, 8] [1, 1]
            : tensor<8x8xf64> into tensor<256x1024xf64>
      }
    }
    %t2 = call @nanoTime() : () -> i64
    %dt = arith.subi %t2, %0 : i64
    return %1, %dt : tensor<256x1024xf64>, i64
  }
}
\end{lstlisting}
}

\end{document}